\documentclass[10pt,journal,compsoc]{IEEEtran}
\usepackage{amssymb}
\usepackage[switch]{lineno}
\usepackage{algorithm}
\usepackage{algorithmic}
\usepackage{booktabs}
\usepackage{multirow}
\usepackage{amsmath}
\usepackage{caption}
\usepackage{graphicx}
\usepackage{diagbox}
\usepackage{subfigure}
\usepackage{stfloats}
\usepackage{array}
\usepackage[table,xcdraw]{xcolor}
\usepackage{url}
\usepackage{float}
\usepackage{color}
\usepackage{stfloats}

\usepackage[flushleft]{threeparttable}

\ifCLASSOPTIONcompsoc
  \usepackage[nocompress]{cite}
\else
  \usepackage{cite}
\fi
\ifCLASSINFOpdf
\else
\fi

\begin{document}
%
\title{Dynamic Virtual Graph Significance Networks for Predicting Influenza}
%
%
%
%

\author{Jie~Zhang,
        Pengfei~Zhou,
        and~Hongyan~Wu
\IEEEcompsocitemizethanks{\IEEEcompsocthanksitem
J.Zhang and P.Zhang are with	the Department of Smart Health, SenseTime, Shanghai, CN, 200233; H.W is with the Joint Engineering Research Center for Health Big Data Intelligent Analysis Technology,
	Shenzhen Institutes of Advanced Technology, Chinese Academy of Sciences, Shenzhen, CN, 518055.\protect\\
E-mail: zhangjie1@sensetime.com; zhoupf@st.btbu.edu.cn; hy.wu@siat.ac.cn}
}

%
%

\markboth{IEEE TRANSACTIONS ON KNOWLEDGE AND DATA ENGINEERING,~Vol.~XX, No.~XX, August~2020}
{IEEE TRANSACTIONS ON KNOWLEDGE AND DATA ENGINEERING,~Vol.~XX, No.~XX, August~2020}

\IEEEtitleabstractindextext{%
\begin{abstract}
  Graph-structured data and their related algorithms have attracted significant
  attention in many fields, such as influenza prediction in public health.
However,  the variable influenza seasonality, occasional pandemics, and domain knowledge pose great challenges to construct an appropriate graph, which could impair strength of the current popular graph-based algorithms  to perform data analysis.  In this study, we develop a novel method,  Dynamic Virtual Graph Significance Networks (DVGSN), which  can supervisedly and dynamically
  learn from similar ``infection situations"  in historical timepoints.
  Representation learning on the dynamic virtual graph can tackle the varied seasonality and pandemics, and therefore improve the performance.
  The extensive experiments on real-world influenza data demonstrate that DVGSN significantly outperforms the current state-of-the-art methods.
  To the best of our knowledge, this is the first attempt to supervisedly learn a dynamic virtual graph for time-series prediction tasks. Moreover, the  proposed method  needs less domain knowledge to build a graph in advance and has rich interpretabilities, which makes the method more acceptable in the fields of public health, life sciences, and so on.

\end{abstract}

\begin{IEEEkeywords}
Representation Learning, Dynamic Virtual Graph,  Influenza Prediction, Time Series.
\end{IEEEkeywords}}

\maketitle

\IEEEdisplaynontitleabstractindextext

%
\IEEEpeerreviewmaketitle

\IEEEraisesectionheading{\section{Introduction}\label{sec:introduction}}

%
%
%
%


\IEEEPARstart{W}{ith} the growing emergence
of graph-structured data such as social networks and biological networks
\cite{lancichinetti2011finding, zachary1977information}, the algorithms to analyze graph data have attracted significant
attention, such as
Graph Convolutional Networks (GCNs) \cite{bruna2013spectral, defferrard2016convolutional, kipf2016semi} and Graph Attention Networks (GAT) \cite{velivckovic2017graph}, etc. The structure of graph data exerts significant impact on the performance of theses   algorithms, because these algorithms heavily depend on the neighborhood relationship of the graph. For example, GNNs iteratively aggregate and integrate the embedding of its neighbors to learn the node embedding of the graph. However, finding out all the influential neighbor nodes and measuring their edge weights appropriately to construct a  graph are nontrivial in many cases, such as in life sciences and public health fields,
which require substantial domain knowledge.

Influenza prediction is an important interdisciplinary problem between computer science and public health.  Influenza circulates worldwide and places a heavy burden on people's health Every year \cite{Influenza2005Key, Brankston2007Transmission}.
The strong infectivity and outbreak of influenza are estimated to result in approximately 35 million cases of symptomatic illnesses, 16 million outpatient medical visits, 490 thousand influenza-associated hospitalizations, and 34 thousand cases of deaths in the influenza season of 2018-2019 in the United States \cite{cdcDiseaseBurden}.
The influenza virus undergoes high mutation rates and frequent genetic re-assortment \cite{LubeckAntigenic, Stech1999Independence, Su1992Heterogeneity}.
To help clinics, hospitals, pharmaceutical companies, and governments better prepare for influenza in a timely manner, we need a reliable model to predict  influenza trends.

\begin{figure}[!htb]
	\centering
	\includegraphics[width=1.01\linewidth]{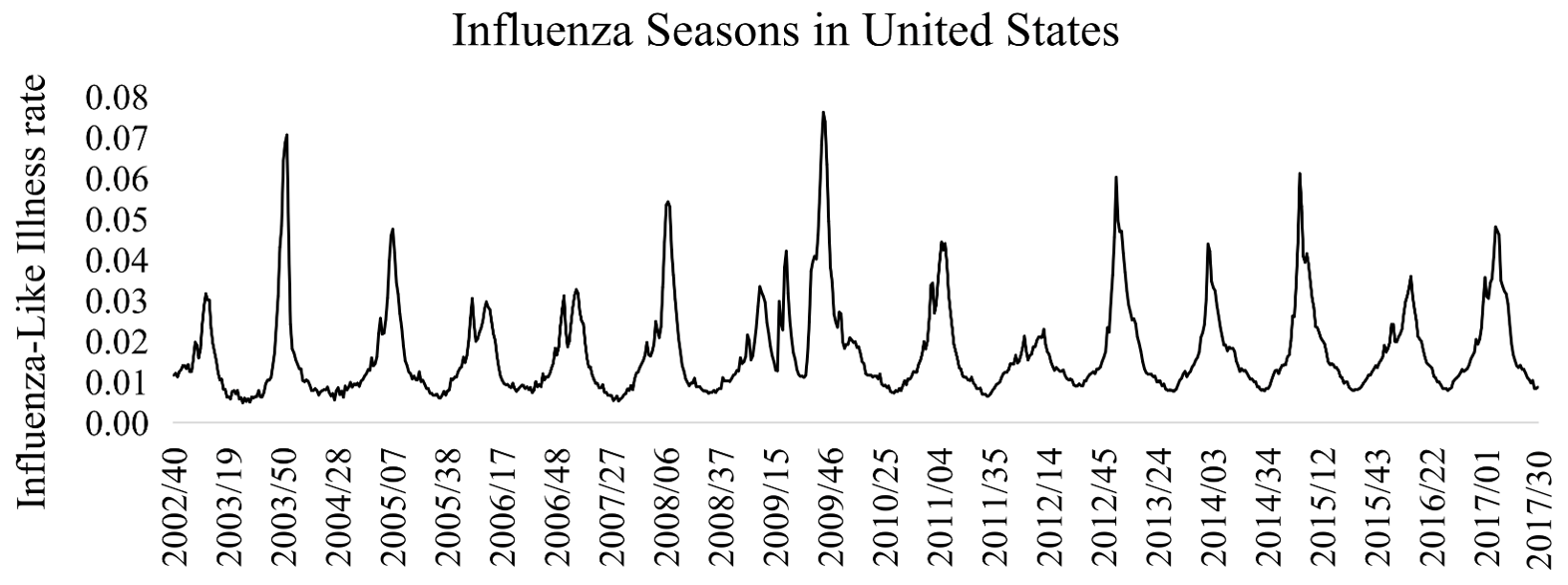}
	\caption{
		The ILI rates of influenza seasons from 2003-2004 to 2016-2017 in the United States. The ILI rate is defined as the number of ILI patients divided by the number of all-illness patients.
	}
	\label{fig:flu_data}
\end{figure}

There are mainly two challenges of predicting influenza.
\textbf{[Challenge 1]}
Influenza seasonality usually varies, from one season to another, in timing, severity, and duration \cite{ PuigFirst,cdc1516summary}.
Table \ref{table:flu_meta_data} shows the descriptive statistics of the Influenza-Like Illness (ILI) rates of influenza seasons from 2003-2004 to 2016-2017 in the United States.
The rates of ``standard deviation / mean" in the ``Highest ILI Rate" and ``Duration" are  33\% and 39\%, respectively.
Such an irregular variation handicaps the predictive methods.
\textbf{[Challenge 2]}
Influenza pandemics occur occasionally but can totally disorder the seasonality for years.
A pandemic is a serious world-wide outburst, resulting from the emerge of a new type of virus and resulting in extremely higher ILI rates, several close and consecutive peaks, and much longer duration, as the piece of the curve around 2009 in Figure \ref{fig:flu_data} shows.
Such a ``mutated" outbreak makes the prediction more difficult.

\begin{table}[htb]

	\caption{
	The columns illustrate the variation of influenza in timing, severity, and lasting, respectively.
		 ``Duration" counts the consecutive weeks, during which the ILI rates are all over 0.01.
	}
		\small
	\label{table:flu_meta_data}
	\begin{center}
		\begin{tabular}{|c|c|c|c|}
			\hline
			Seasons                                                           & \begin{tabular}[c]{@{}c@{}}Peak Week \\ (year/week)\end{tabular} & \begin{tabular}[c]{@{}c@{}}Highest \\ ILI Rate\end{tabular} & \begin{tabular}[c]{@{}c@{}}Duration \\ (weeks)\end{tabular} \\ \hline
			2003-2004                                                         & 2003/06                                                          & 0.0317                                                      & 28                                                          \\ \hline
			2004-2005                                                         & 2003/52                                                          & 0.0706                                                      & 25                                                          \\ \hline
			2005-2006                                                         & 2005/07                                                          & 0.0475                                                      & 37                                                          \\ \hline
			2006-2007                                                         & 2005/52                                                          & 0.0305                                                      & 31                                                          \\ \hline
			2007-2008                                                         & 2007/07                                                          & 0.0327                                                      & 31                                                          \\ \hline
			2008-2009                                                         & 2008/07                                                          & 0.0542                                                      & 31                                                          \\ \hline
			\multirow{3}{*}{2009-2010}                                        & 2009/06                                                          & 0.0334                                                      & \multirow{3}{*}{91}                                         \\ \cline{2-3}
			& 2009/21                                                          & 0.0421                                                      &                                                             \\ \cline{2-3}
			& 2009/42                                                          & 0.0762                                                      &                                                             \\ \hline
			2010-2011                                                         & 2011/05                                                          & 0.0444                                                      & 37                                                          \\ \hline
			2011-2012                                                         & 2012/11                                                          & 0.0229                                                      & 41                                                          \\ \hline
			2012-2013                                                         & 2012/52                                                          & 0.0603                                                      & 44                                                          \\ \hline
			2013-2014                                                         & 2013/52                                                          & 0.0439                                                      & 43                                                          \\ \hline
			2014-2015                                                         & 2014/52                                                          & 0.0611                                                      & 41                                                          \\ \hline
			2015-2016                                                         & 2016/10                                                          & 0.0359                                                      & 42                                                          \\ \hline
			2016-2017                                                         & 2017/06                                                          & 0.0481                                                      & 44                                                          \\ \hline
			MEAN                                                              & -                                                                & 0.0460                                                      & 40                                                          \\ \hline
			\begin{tabular}[c]{@{}c@{}}standard deviation\\ (SD)\end{tabular} & -                                                                & 0.0152                                                      & 16                                                          \\ \hline
			SD/MEAN                                                           & -                                                              & 33\%                                                        & 39\%                                                        \\ \hline
		\end{tabular}
	\end{center}
\end{table}

The existing machine/deep learning models, such as
XGBoost (XGB),
Temporal Pattern Attention Long Short-Term Memory (TPA-LSTM),
Temporal Convolutional Networks (TCN), and
Transformer,
use current and historical values  in a user-defined time window as input to predict future values.
These methods lack considering similarities outside the time window.
Although one can simply increase the length of the time window to include more information,
there are always some timepoints outside the window.
Besides, the bigger the length of the time window is, the fewer the training instances will be left, which makes the predictive model unreliable.
If a method can dynamically find historical timepoints that have similar ``infection situations" as auxiliary information,
the model could tackle the varied seasonality and the occasional pandemics.

However, how to accurately represent the ``infection situations" poses a challenge since the situation should include the information of the influenza severity, the tendency, the duration and other factors that  may be beyond our knowledge.



\begin{figure}[!htb]
	\centering
	\includegraphics[width=1\linewidth]{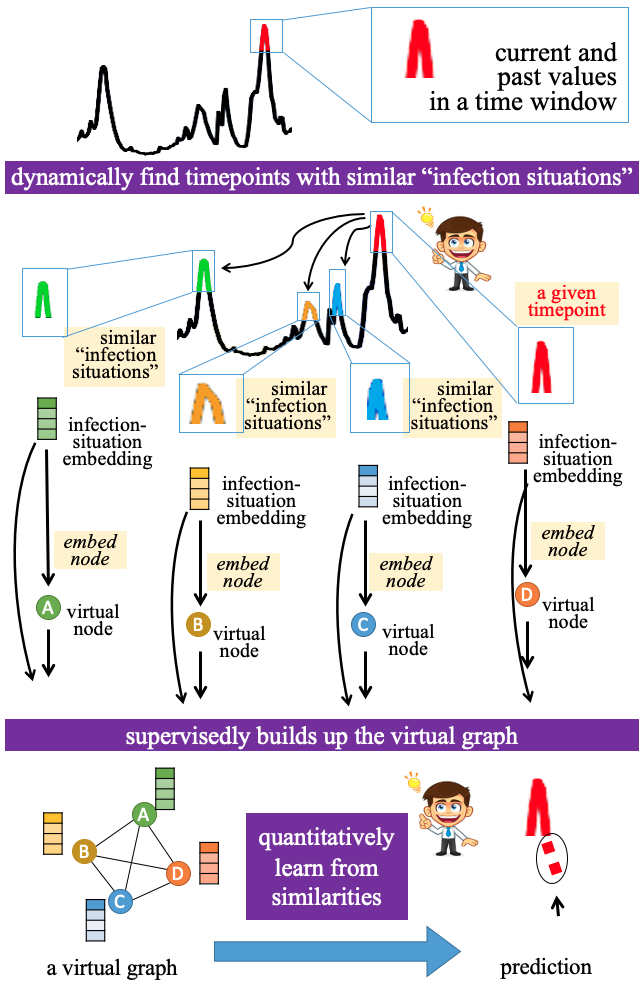}
	\caption{
		The architechture of the proposed method---DVGSN.
	}
	\label{fig:idea}
\end{figure}

In this study, we develop a novel method, namely Dynamic Virtual Graph Significance Networks (DVGSN), as Figure \ref{fig:idea} illustrates.
DVGSN constructs a virtual graph for influenza prediction.
In the virtual graph, a virtual node represents a timepoint.
The embedding of a virtual node represents the ``infection situations" at the timepoint.
The virtual edges connect two virtual nodes at two timepoints, and the edge weights measure the significance of the virtual edge.
Since the timepoints connected by the virtual edges can be outside the time window,
DVGSN can break the limitation of the time window by learning from neighbor nodes and improve the predictive accuracy.

A natural static graph defined with domain knowledge beforehand could not align well with the specific analytical task.
As a result, the ``neighborhood" in an ``unsupervised graph" could be improper for the specific analytical task and damage the analytical outcomes.
Different from a natural graph with static nodes and edges,
in a dynamic virtual graph, every node and edge are supervisedly  dynamically learned during the training procedure in the prediction task.
Moreover, a virtual graph naturally has rich interpretabilities.
For example, similar ``infection situations" found by the virtual graph can provide us with clues how the virtual graph finds similarities and how the proposed method performs the prediction for pandemics.
The interpretabilities make the proposed method more acceptable, especially in the fields of epidemiology and public health, in which researches usually emphasize the interpretabilities of the predictive models for further government measures, etc.

The contributions of this work are concluded as follows.

(1) To the best of our knowledge, this is the first attempt to supervisedly learn a dynamic virtual graph for time-series prediction.

(2) The proposed method  need less domain knowledge to build a graph in advance and has rich interpretabilities, which are indispensable in epidemiology, public health, and the like.

(3) We carry out extensive experiments on the real-world data, and the experimental results prove that the proposed method significantly outperforms the state-of-the-art methods.


\section{RELATED WORK}

This section describes the previous work  from the point of view of influenza prediction and graph-based deep learning.


\subsection{Influenza Prediction}
The machine/deep learning for forecasting influenza or other time-series data are mainly  categorized into two groups.
Firstly, some researchers focus on looking for effective ``features".
For example, search engine query data are used for prediction influenza in \textit{Google Flu Trends}\footnote{https://www.google.org/flutrend
} \cite{GINSBERG2009Detecting, Lee2017Forecasting}.
Twitter data are also used in other research papers \cite{MolaeiPredicting, Li2013Early}.
However, these models usually suffer from the unreliable source of huge amounts of information from such as  internet searches. For example, Google’s algorithm was quite vulnerable to overfitting to seasonal terms unrelated to the flu, like “high school basketball”. This example also demonstrates the importance of model interpretability.
Secondly, other researchers focus on looking for effective ``models", such as
RF \cite{darwish2020comparative, KaneComparison, zhang2017comparative},
Gradient Boosting \cite{darwish2020comparative, zhang2017comparative},
Multilayer Perceptron (MLP) \cite{darwish2020comparative, zhang2017comparative},
Long Short Term Memory (LSTM) \cite{darwish2020comparative, zhang2017comparative, yin2020tempel},
Transformer (TFR) \cite{wu2020deep},
and so on. Deep learning based methods, e.g. Transformer, are drawing more attention  for  their accuracy while most of them suffers from the poor interpretability.

Moreover, statistical models and dynamic analysis models are considered easily accessible tools for simulating patterns of infection by influenza, such as SI, SIS, SIR model and their variants \cite{Dukic2012Tracking}. However, their parameters are  subject to change and the approximation of the parameters is difficult \cite{WuInfluence}, such as the basic reproduction number $R_0$, population mobility etc.

\subsection{Graph-based deep learning} \label{rl:gnn}
For mining a natural graph, such as
Cora \cite{sen2008collective} and Digg \cite{hogg2012social},
Graph Neural Networks (GNNs) are usually used, such as
GCN, GAT, and
Graph Isomorphism Network (GIN) \cite{xu2018powerful}.
In an analytical task without a natural graph, to leverage powerful GNNs, a  graph can be  constructed beforehand.
Researches need to use domain knowledge, such as
medicine and
transportation \cite{geng2019spatiotemporal},
and mathematical calculation, such as
Euclidean distance \cite{fu2018comparison}, to construct a  graph beforehand.
Nonetheless, all of these graphs  are thought of as ``unsupervised graphs"
because the calculation for the construction is not updated by backpropagation for the specific analytical task.
In other words, an ``unsupervised graph" could NOT align with the specific analytical task.
As a result, the ``neighborhood" in an ``unsupervised graph" could be improper for the specific analytical task and damage the analytical outcomes.
In this study,
we  develop a method to construct a ``supervised graph",
which could dynamically and supervisedly learn the effective information from other instances during the training procedure in the specific analytical task.


\section{THE PROPOSED MEHTOD}
\subsection{Influenza prediction tasks}
We formally define the prediction task with  the classic machine/deep learning algorithms  as Formula  \ref{formula:multistep}:
\begin{equation}
\label{formula:multistep}
\begin{split}
\boldsymbol{\hat{y}_{(v,q)}} =& [\hat{y}_{(v+1)}, \hat{y}_{(v+2)}, \ldots , \hat{y}_{(v+q)}]^T \\
=& [f_1(O_{(v,p)}), f_2(O_{(v,p)}), \ldots, f_q(O_{(v,p)})]^T \\
\end{split}
\end{equation}
where
$\boldsymbol{\hat{y}_{(v, q)}} \in \mathbb{R}^q$ is the vector to be predicted,
$v$ is a given timepoint, and
$q$ is the predictive window size;
$\hat{y}_{(v+i)}$ is the predicted value of the upcoming $i$-th week,
and $f_i(\cdot)$ is a time-series model to predict the value of the upcoming $i$-th week;
$O_{(v,p)} = [o_v, o_{(v-1)}, \ldots, o_{(v-p+1)}]$ is the observed time-series values with a time lag $p$,
$o_v$ is the observed value, and
$o_{(v-i)}$ is the value of the past $i$-th week.

There are  two types of time-series prediction:
(a) single-step influenza prediction and
(b) multi-step prediction.
A single-step prediction predicts the value for one step in advance ($q=1$ in Formula \ref{formula:multistep} ),
and a multi-step prediction predicts the consecutive values with a bigger predictive window size ($q>1$ in Formula \ref{formula:multistep}).

As Formula  \ref{formula:multistep} shows, the classic methods heavily depends on the observations in the time window but lacks considering historical similarities outside the time window.
Table \ref{tab:notation} presents the notations utilized in this work.

\begin{table}[htbp]
	\caption{\label{tab:notation}Notations and Explanations.}
	\label{tab:HIN}
	\small
	\begin{tabular}{cl}

		\toprule
		Notations & Explanations\\
		\midrule

		$O$ & the observed time-seies data \\
		$p$ & the time lag \\
		$q$ & the predictive window size \\

		$\boldsymbol{y}_{(v,q)}$ & \multicolumn{1}{l}{\begin{tabular}[l]{@{}l@{}} the vector of the true values of the predictive \\ window size of $q$  for the node $v$ in the virtual graph \end{tabular}}\\
		$\boldsymbol{\hat{y}}_{(v,q)}$ & \multicolumn{1}{l}{\begin{tabular}[l]{@{}l@{}} the vector of the predicted values of the predictive \\ window size of $q$  for the node $v$ in the virtual graph \end{tabular}}\\

		$\mathcal{G}$ & the virtual graph \\
		$\mathcal{V}$ & the set of all virtual nodes \\
		$\mathcal{E}$ & the set of all virtual edges \\

		$\chi$ & the observed matrix of the ILI rates \\
        $S$ & the node embedding matrix of the virtual graph \\
        $T$ & the adjacency matrix of the virtual graph \\

		\bottomrule

	\end{tabular}
\end{table}



\subsection{Dynamic Virtual Graph}

As aforementioned in Table 1 and Figure 1,
Influenza seasons that vary in timing, severity, and duration.
And pandemics mutate the influenza outbreaks.
Dynamically looking for similar ``infection situations"  instead of sticking to a fixed periodicity (roughly one year) could be a key to varied seasonality and pandemics for influenza prediction. We formally define the concept of a dynamic virtual graph.

\noindent\textbf{Dynamic Virtual Graph}. Different from a natural graph with static nodes and edges,
we define a  virtual graph as  $\mathcal{G} = (\mathcal{V}, \mathcal{E})$
with a set of nodes ($\mathcal{V}$)
and a set of edges ($\mathcal{E}$), in which every node and edge are supervisedly or semi-supervisedly  dynamically learned during  the training procedure in the prediction task.

\begin{figure}[!htb]
	\centering
	\includegraphics[width=1\linewidth]{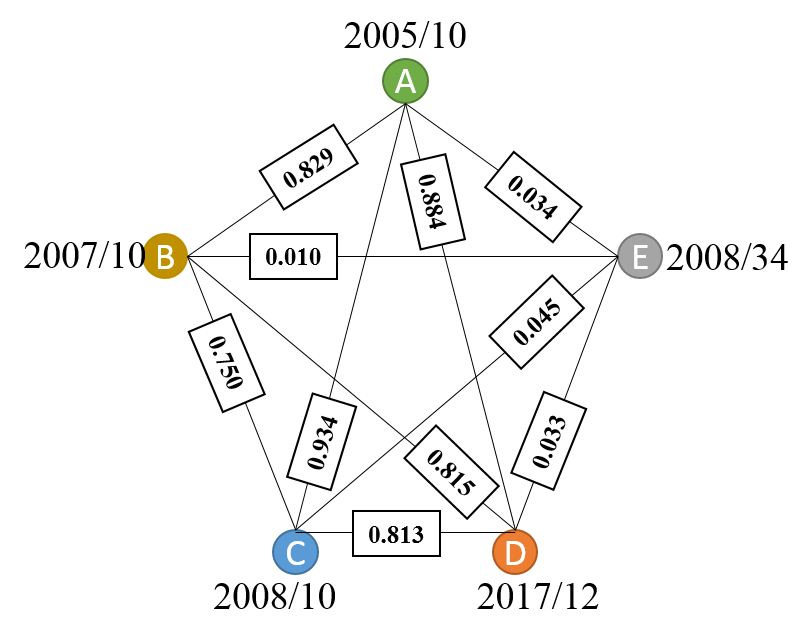}
	\caption{
		An image of a virtual graph.
		A virtual node represents a timepoint in the format of ``year/week".
		A virtual edge connects two virtual nodes  at two timepoints.
		The values in the boxes above the edges are the significance of the virtual edges. }
	\label{figure:virtual-graph}
\end{figure}

In this study each node is a function that can be trained to capture ``infection situations", and each edge describes the significance of the similarity. Figure \ref{figure:virtual-graph} gives an image of a virtual graph to predict influenza in this study.
A virtual node $v$ ($v  \in \mathcal{V}$) represents a timepoint, such as 2017/12, which means the 12th week in 2017.
The embedding of the virtual node, representing the comprehensive ``infection situations" at the timepoint, is denoted as $s_v$ for the node $v$.
A virtual edge $e$ ($e  \in \mathcal{E}$) connects two virtual nodes at two timepoints, and the edge weights measure the significance of the virtual edge.

\subsection{Virtual Node Representation}

The virtual node representation will be used (1) to perform ``infection-situations" embedding for the subsequent representation learning and (2) to learn the significance of the similarity among the different ``infection-situations".
How to define a proper function of the virtual node representation vector $s_v$ for the node $v$ is a pivotal problem.
An ``infection-situation" embedding vector needs to include comprehensive infective information, such as :

(a) the timing, severity, and duration of the infection at a given timepoint;

(b) the first-order differences (``speed") and the second-order differences (``acceleration") at a given timepoint;

(c) tendency (upward, downward, fluctuation, or a turning point) at a given timepoint;

(d) descriptive statistics (mean, median, maximum, minimum, variance, and the like) at a given timepoint.

Many previous researches studied how to ``unsupervisedly" represent ``situation" for prediction  \cite{alberg2017improving, crone2016feature, frank2001time, ghiassi2005dynamic}.
Some use current and past values in the time-series data as input.
Others use domain knowledge, such as
varying lag structures at different steps \cite{crone2016feature},
to present ``situation" vectors.
Nonetheless, these methods of  ``unsupervised presenting" is separate from the model training.
In this study, we propose a supervised representation of ``situation" to learn a more appropriate presentation for a given analytical task. Formula \ref{formula:sv} illustrates the node representation of ``infection-situations":
\begin{equation}
\label{formula:sv}
\begin{split}
s_v & =  \tau(obs_v, obs_{(v-1)}, ... \mbox{ }, obs_{(v-p+1)}; \theta_{\tau}) \\
&= \tau(O_{(v, p)}; \theta_{\tau}) \\
& = \sigma(W_{MLP-2}(\sigma(W_{MLP-1} O_{(v, p)})))
\end{split}
\end{equation}
where
$\tau$ is a neural network,
$\theta_{\tau}$ is the parameter for ($\tau$), and
$\sigma$ is an activation function for which we use Exponential Linear Units (ELU) in this work.
 In this work, a two-layered Multilayer Perceptron (MLP) is adopted.
$W_{MLP-1} \in \mathbb{R}^{p \times 256} $  and $W_{MLP-2}  \in \mathbb{R}^{256 \times 256} $  are the trainable weights of the first and second layer of MLP, respectively.

There are three reasons why we need to project the original feature space nonlinearly into a high-dimensional space as an embedding vector of a virtual node:

\textbf{(a) To work as time-series feature extraction.}

Time-series analyses usually adopt arbitrary feature engineering, such as Kalman Filtering \cite{Links1995An}, and so on.
How to select effective feature engineering is non-trivial, usually experienced-based and time-consuming.
Since an MLP can theoretically simulate any function \cite{Hornic1989Multilayer, LeshnoMultilayer,HornikApproximation},
implementing MLP can perform effective representation in a high-dimensional space.

\textbf{(b) To supervisedly and dynamically learn virtual node representations.}

Since the MLP is a part of the entire end-to-end learning,
the projected embedding vectors are supervisedly learned from and for a specific analytical task.
Such a supervisedly-learned embedding space is supposed to work better than the original feature space, which is static and cannot be updated or learned, and thereby improve the accuracy.

\textbf{(c) To project virtual node embedding to an appropriate space.}

The high-dimensional embedding vectors of two virtual nodes will be used to define the significance of a virtual edge.
A high-dimensional space that represents a variety of complex time-series characteristics can work better than the original feature space that just consists of the ILI rates of current and past few weeks.

\subsection{Virtual Edge Significance } \label{Sec:ve}

To a given node, different neighbors may have different similarities in ``infection situations".
The significance of a virtual edge needs to be decided.
In this study, we measure the significance of the virtual edge between the node $u$ and $v$ by performing inner product on the after linear projection and instance normalization on the high-dimensional embedding vectors of ``infection-situations", as Formula \ref{formula:similarity} illustrates:

\begin{equation}
\label{formula:similarity}
\begin{aligned}
\begin{split}
t_{(v, u)} &= \kappa(s_{v}, s_{u}) \\
&=  inst\_norm[W_{line\_proj} s_{v}] \odot inst\_norm[W_{line\_proj} s_{u}]  \\
\end{split}
\end{aligned}
\end{equation}
where
$t_{(v, u)}$ is the the significance of the virtual edge from the node $u$ to $v$,
$inst\_norm(\cdot)$ is the instance normalization,
$W_{line\_proj}  \in \mathbb{R}^{256 \times 256} $ is the trainable weight of the linear projection and
$\odot$ is the inner product.

The significance  ($t_{(v, u)}$) of the virtual edge from the node $u$ to $v$ has some properties:

\textbf{(a)}
\textbf{$-1 \le t_{(v, u)} \le 1$}

The value range of virtual edge significance is $[-1,1]$.

\textbf{(b)}
\textbf{$ t_{(v,v)} = 1$}

The significance of the self-loop virtual edge  is ``1".

\textbf{(c)}
\textbf{$t_{(v, u)} = t_{(u, v)}$}

The virtual edge significance is symmetric.

\subsection{Graph Significance Networks} \label{Sec:vGNN}

The virtual graph, which is composed of the virtual nodes representing the ``infection situation" at each timepoint and the virtual edges with the similarity significance,  is input into  GNNs.
For a given node, the GNNs iteratively aggregate and integrate the embedding of its neighbors to learn a representation vector ($\boldsymbol{h}_v$).
Formula \ref{formula:normal_GNN} illustrates the $l$-th iteration of aggregation and integration:
\begin{equation}
\label{formula:normal_GNN}
\begin{aligned}
\begin{split}
\boldsymbol{h}^{(l)}_{v} & =INT( \boldsymbol{h}^{(l-1)}_{v}, AGG(\boldsymbol{h}^{(l-1)}_{u};  \forall u \in \mathcal{N}(v)))
\end{split}
\end{aligned}
\end{equation}
where
$\boldsymbol{h}^{(l)}_{v}$ and $\boldsymbol{h}^{(l)}_{u}$ is the representation vector at the $l$-th iteration/layer of the given node $v$ and the neighbor node $u$ , respectively;
$\mathcal{N}(v)$ is the set of neighbor nodes;
$AGG(\cdot)$ is a function that aggregates the infective information from its neighbors $\mathcal{N}(v)$ ---the timepoints that have similar ``infection situations", and
$INT(\cdot)$  is a function that integrates the infective information from the given timepoint $v$ itself and the aggregated infective information by $AGG(\cdot)$ based on its neighbors $\mathcal{N}(v)$.

A variety of functions of $AGG(\cdot)$  and $INT(\cdot)$ have been proposed in the previous studies \cite{velivckovic2017graph, xu2018powerful}.
In this work, we initiate and update the node representations  as Formula \ref{formula:iteration} shows:
\begin{equation}
\label{formula:iteration}
\begin{aligned}
& \boldsymbol{h}^{(0)}_{v} = \boldsymbol{s}_v \\
& \boldsymbol{h}^{(l)}_{v} = \sigma(W_{GNN-l} \cdot \sum\nolimits_{u \in \mathcal{N}(v) \cup {\{v\}}} (t_{(v, u)} \cdot \boldsymbol{h}^{(l-1)}_{u} )) \\
\end{aligned}
\end{equation}
where
$W_{GNN-l} \in \mathbb{R}^{256 \times 256} $  is the trainable weight of the l-th layer of GNNs. In this work, we adopt 2-layered GNNs.

\subsection{Regressive Layer}

The sum of the initial virtual node representation ($\boldsymbol{h^{0}_{v}}$), the representation vector after the first GNN layer ($\boldsymbol{h^{1}_{v}}$), and the second GNN layer ($\boldsymbol{h^{2}_{v}}$) is input into a regressive layer (implemented by a linear layer) to achieve the final prediction, as Formula \ref{formula:regressive} shows:
\begin{equation}
\label{formula:regressive}
\begin{aligned}
\begin{split}
\boldsymbol{\hat{y}}_{(v,q)} = W_{regr} (\boldsymbol{h^{0}_{v}} + \boldsymbol{h^{1}_{v}} + \boldsymbol{h^{2}_{v}}) \\
\end{split}
\end{aligned}
\end{equation}
where
$W_{regr} \in \mathbb{R}^{256 \times q} $  is the trainable weight of the final regressive layer.

\subsection{Loss Function}

The loss function is defined as Formula \ref{eq:loss} shows:
\begin{equation}
\label{eq:loss}
\mathcal{L} = \frac{1}{n \times q}\sum_{v \in \mathcal{V}} || \boldsymbol{y}_{(v,q)} - \boldsymbol{\hat{y}}_{(v,q)} ||_2^2 + \lambda || \boldsymbol{T}||_F^2
\end{equation}
where $\frac{1}{n \times q}\sum_{v \in \mathcal{V}} || \boldsymbol{y}_{(v,q)} - \boldsymbol{\hat{y}}_{(v,q)} ||^2 $ is the predictive loss in Mean Square Error (MSE)
($n$ is the number of virtual nodes),
$\boldsymbol{y}_{(v,p)}$ and $\boldsymbol{\hat{y}}_{(v,p)}$  are the vectors of the true and predicted values, respectively.
$\boldsymbol{T}$ is the adjacency matrix of the virtual graph, and $|| \boldsymbol{T}||$ is the penalty term to
limit the complexity of the virtual graphs and
improve the robustness of the model, and $|| \cdot ||_F$ represents the matrix Frobenius norm); and
$\lambda$ is an adjustable hyper-parameter to balance the two parts of losses.

\begin{figure}[!htb]
	\centering
	\includegraphics[width=.8\linewidth]{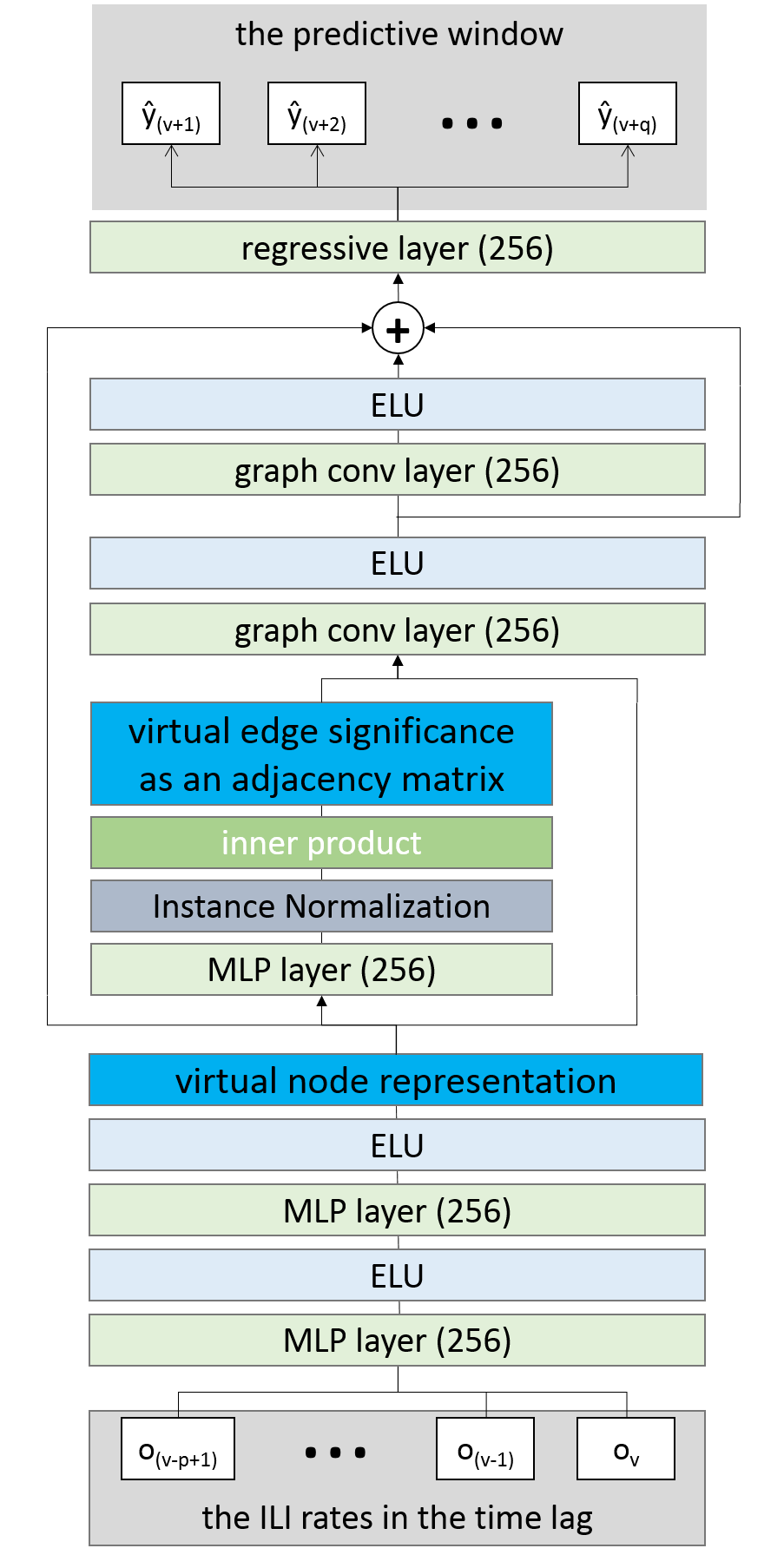}
	\caption{The structure of the proposed DVGSN.}
	\label{figure:residual-gnn}
\end{figure}

The entire algorithm of the proposed methods is shown in Figure \ref{figure:residual-gnn} and Algorithm \ref{algorithm:residual_virtual_gnn}.

\begin{algorithm}[!htb]
	\renewcommand{\algorithmicrequire}{\textbf{Input:}}
	\renewcommand{\algorithmicensure}{\textbf{Output:}}
	\caption{The proposed DVGSN.}
	\label{algorithm:residual_virtual_gnn}
	\small

	\begin{algorithmic}[1]

		\REQUIRE~~\\
		The observed time-series data: $O$, \\
		The time lag: $p$, \\
		The predictive window size: $q$, \\
		The training epochs: $I$, \\
		The adjustable hyperparameter of levels: $\lambda$. \\

		\ENSURE~~\\
		The predictive model for influenza in United States.

		\STATE Prepare the observed matrix: $ \chi= [O_{(v,p)}]$, $\forall v \in \mathcal{V}$;
		\STATE Prepare the target matrix: $Y = [\boldsymbol{y}_{(v,q)}]$, $\forall v \in \mathcal{V}$;
		\STATE Calculate the number of virtual nodes: $n = |O| - p - q +1$;

		\FOR{$i = 1, 2, \cdots,  I$}

			\STATE $S  \leftarrow elu(1d\_conv(elu(1d\_conv(\chi))));$
			\STATE $T \leftarrow inst\_norm(1d\_conv(S)) (inst\_norm(1d\_conv(S)) )^{T};$
		    \STATE $H^{0}  \leftarrow S$;
		    \STATE $H^{1}  \leftarrow  GNN(S,T)$;
		    \STATE $H^{2}  \leftarrow  GNN(H^{1},T)$;
		    \STATE $\hat{Y} = regressive\_layer ({H^{0}} + {H^{1}} + {H^{2}}) $;
			\STATE $ \mathcal{L}  = || Y- \hat{Y} ||_F^2 + \lambda  \times || T ||_F^2; $
			\STATE Perform back propagation and update parameters;

		\ENDFOR

	\end{algorithmic}
\end{algorithm}

\subsection{Comparison with the existing methods}

\textbf{(a) The limitation of time window }

The prediction for a given timepoint $v$ by DVGSN can be simplified and formularized as Formula \ref{eq:entire}:
\begin{equation}
\label{eq:entire}
\begin{aligned}
\begin{split}
\hat{y}_{(v,q)}  &= f_\varsigma (s_{v}, s_{u}, \mathcal{E}; \theta_{\varsigma})  \\
& = f_\varsigma (s_{v}, s_{u}, \kappa(s_{v}, s_{u}) ; \theta_{\varsigma}) \\
& = f_\varsigma (\tau(O_{(v, p)}), \tau(O_{(u, p)}), \kappa(\tau(O_{(v, p)}), \tau(O_{(u, p)})); \theta_{\tau},\theta_{\varsigma})\\
& = f_\varsigma (\tau(O), \kappa (\tau(O)); \theta_{\tau},\theta_{\varsigma}) \\
& = f_\varsigma (O; \theta) \\
\end{split}
\end{aligned}
\end{equation}
where
$f_\varsigma $ is the algorithm of DVGSN and
$\theta$ represents the set of all the parameters in DVGSN.

To perform prediction for the timepoint $v$, all the observed time-series data ($O$) in the training dataset are input into  DVGSN.
Comparatively, the input in the classic machine/deep learning methods (as formula \ref{formula:multistep} shows) are just the observed values ($O_{(v, p)}$) in the user-defined time window with the time lag ($p$).
Inputting all the observed time-series data ($O$) can   capture similar ``infection situations" from all the timepoints instead of sticking to the fixed static nodes and edges.

\textbf{(b) Other differences}

    The virtual nodes and virtual edges in  DVGSN are supervisedly learned during the training procedure in the specific prediction task while the other existing GNNs-based methods use a static  graph defined beforehand. Another difference lies in the $AGG(\cdot)$ function in Formula \ref{formula:normal_GNN}.
The algorithm of the $l$-th layer of iteration in the  attentive GNNs (such as GAT) and DVGSN is illustrated as  Formula  \ref{formula:normal_GAT} and Formula  \ref{formula:normal_GSN_AGG} shows, respectively:
\begin{equation}
\label{formula:normal_GAT}
\begin{aligned}
\begin{split}
\boldsymbol{h^{(l)}_{v}} & = \sigma(W \cdot \Sigma_{u \in \mathcal{N}(v)\cup {\{v\}}} (\alpha_{(v, u)} \cdot \boldsymbol{h^{(l-1)}_{u}}) \\
\alpha_{(v, u)}  &= \frac{exp(\sigma (a^{T} [W\boldsymbol{h_v^{(l-1)} }|| W\boldsymbol{h_u^{(l-1)}}]))}{\Sigma_{k \in \mathcal{N}(v)}exp(\sigma (a^{T}  [W\boldsymbol{h_v^{(l-1)}} || W\boldsymbol{h_k^{(l-1)}}])}  \\
&, \forall  u \in \mathcal{N}(v) \\
\end{split}
\end{aligned}
\end{equation}

where
$\alpha_{(v, u)}$ is the normalized attention coefficient,
$(\cdot)^{T}$ represents matrix transposition, and
$||$ is the concatenation operation.

\begin{equation}
\label{formula:normal_GSN_AGG}
\begin{aligned}
\begin{split}
 \boldsymbol{h}^{(l)}_{v} & = \sigma(W_{GNN-l} \cdot \sum\nolimits_{u \in \mathcal{N}(v) \cup {\{v\}}} (t_{(v, u)} \cdot \boldsymbol{h}^{(l-1)}_{u} )) \\
t_{(v, u)} &=  inst\_norm(W_{line\_proj} s_{v}) \odot inst\_norm(W_{line\_proj} s_{u})  \\
&, \forall  u \in \mathcal{N}(v) \\
\end{split}
\end{aligned}
\end{equation}


In  GAT,
the input  ($\alpha_{(v, u)} \cdot \boldsymbol{h^{(l-1)}_{u}}$)
is a $mean$ (precisely a weighted $mean$) of the embedding vectors in the neighborhood since $\sum_{u \in \mathcal{N}(v)} \alpha_{(v, u)} = 1$ regardless of the graph structure.
Comparatively, in  DVGSN, the aggregation function $\sum(t_{(v, u)} \cdot \boldsymbol{h^{(l-1)}_{u}})$ is a $sum$ of the embedding vectors in the neighborhood on condition that $-1 \le t_{(v, u)} \le 1$ holds.
Theoretically, the expressive power of $mean$ based aggregators is weaker than $sum$ aggregators
because $sum$ captures the full multiset while $mean$ captures the proportion / distribution of elements of a given type \cite{xu2018powerful}.


\section{EXPERIMENTS}

\subsection{Data}

We scrape the influenza data of the United States from 2003/30 to 2017/30 in the ``FluView Interactive" \cite{cdcFluview},
a website of Centers for Disease Control and Prevention, National Center for Immunization and Respiratory Diseases.
The weekly ILI rates are calculated and used for this work.
Figure \ref{fig:flu_data} illustrates the time-series plot of the ILI rates.
The piece of the curve around 2009, which has three consecutive peaks, is a pandemic in 2009.
Table \ref{table:flu_meta_data} summaries the descriptive statistics of the influenza seasons from 2002-2003 to 2016-2017.
The ``mean $\pm$ standard deviation" of the column of ``The Highest ILI Rate" is $0.0460 \pm 0.0152$.
Besides, the standard deviation ($0.0152$) is around 33\% higher than the value of  mean ($0.046$), presenting a considerable variance in severity.
The ``mean $\pm$ standard deviation" of the column of ``Duration" is $40 \pm 16$.
The standard deviation ($16$) is around 40\%h of the mean ($40$), presenting a considerable variance in lasting.
Moreover, the column of ``The Peak week" demonstrates the timing of influenza seasons  varies year by year.

\subsection{Baseline}

In this work, the baseline models include a variety of state-of-the-art models. We do not compare with the SI-based models because it is difficult to obtain the values of the parameters such as the number of susceptible and infected individuals.

\textbf{$\bullet$ Autoregression (AR)}.
An AR model is a statistical method that uses observed values from current and past time steps as input and  implements a linear regression to predict future values.

\textbf{$\bullet$ $\mathnormal{k}$-Nearest Neighbors($\mathnormal{k}$-NN)}.
The regression of $\mathnormal{k}$-NN examines the values of a chosen number ($\mathnormal{k}$) of data points surrounding a target data point, and uses the mean of the values as prediction.
The $\mathnormal{k}$-NN regression can be used for time-series prediction \cite{mehdizadeh2020using}.

\textbf{$\bullet$ Random Forest(RF)}.
A RF model is an ensemble of decision trees trained with the “bagging” method, which leverages a combination of learning models and thereby increases the overall performance \cite{wu2017time}.

\textbf{$\bullet$ XGB}.
The XGB regression implements the framework of Gradient Boosting by providing a parallel boosting, which consists of iteratively learning weak regressors with respect to a distribution and adding them to a final strong regressor \cite{abbasi2019short}.

\textbf{$\bullet$ Multilayer Perceptron (MLP)}.
An MLP is a neural network, in which each node in a layer is fully connected to every node in the adjacent layers and fit a non-linear function.
The MLP can be used for time series analyses by mapping current and past values to one or multiple future predictive values \cite{cao2019financial}.


\textbf{$\bullet$ TPA-LSTM}.
The TPA-LSTM uses a set of filters to extract time-invariant temporal patterns. The extraction is similar to transforming time series data into its “frequency domain” for forecasting \cite{shih2019temporal}.

\textbf{$\bullet$ TCN}.
A TCN uses a causal and dilated convolutional network to predict sequential data \cite{bai2018empirical}, such as time series \cite{hewage2020temporal}, and so on .

\textbf{$\bullet$ TFR}.
The TFR is a model relying entirely on attention mechanism to compute representations of its input and output without using sequence aligned RNNs or convolution \cite{NIPS2017_7181}. TFR can also be used for time -series prediction \cite{wang2019effects}.

\textbf{$\bullet$ GAT}.
The GAT algorithm is a type of GNNs that considers the attention mechanism on graphs \cite{velivckovic2017graph}.

\textbf{$\bullet$ GIN}.
The GIN model is a theoretically designed model for analyzing the expressive power of GNNs to capture different graph structures. The GIN is provably the most expressive among the class of GNNs and is theoretically as powerful as the Weisfeiler-Lehman graph isomorphism test \cite{xu2018powerful}.

\subsection{Implementation Details}

For all the baseline models and the proposed model, we randomly initialize parameters with the uniform distribution and select the Adam optimizer\cite{kingma2014adam} with a learning rate of 0.001.
We set all hidden layer size of 256 units and use the activation function of an ELU.
We set the epoch to 200 and choose the parameter with the best result on the validation set.
To ensure fairness, we split the datasets and use the same
training set (the current timepoints are from 2003/41 to 2012/02),
validation set (the current timepoints are from 2012/03 to 2014/42), and
testing set (the current timepoints are from 2014/43 to 2017/29 when the predictive window size is 1;
the current timepoints are from 2014/43 to 2017/27 when the predictive window size is 3;
the current timepoints are from 2014/43 to 2017/24 when the predictive window size is 6) for all the models in this work.
We set all the models to the same depth.
For MLP,
we use a five-layer structure, which contains an input layer, four hidden layers, and an output layer (regressive layer).
For TPA-LSTM,
we feed the feature vectors into a TPA-LSTM layer, then input the hidden state to 3-layered MLP, and finally implement an output layer.
For TFR,
we use two TFR blocks, which correspond to four layers.
For GAT,
we  use four layers of GAT and one output layer.
For the proposed method, in each backpropagation, we use all training data as model input and randomly select batch nodes from the training set to calculate the loss and update model parameters.  In this work, to use the algorithm of GIN and GAT, we also connect a given virtual node to all the other virtual nodes, as DVGSN does.
Our source code and dataset are available at https://github.com/aI-area/DVGSN.

\subsection{Results}

\begin{table*}[t]
	\caption{The MSEs in all the models. The bold font indicates the best performance in each test.}
	\label{table:result-1}
	\begin{center}
	\resizebox{\textwidth}{!}{
		\begin{tabular}{|c|c|ccccccccccc|}
			\hline
		\rowcolor[HTML]{c7c9cc}
			p $^{\rm *}$  &
			q $^{\rm *}$ &
			AR &
			kNN &
			RF &
			XGB &
			MLP &
			TPA-LSTM &
			TCN &
			TRF &
			GAT &
			GIN &
			DVGSN \\ \hline
			\multirow{3}{*}{6}  & 1 & 0.0796 & 0.1177 & 0.0897 & 0.1071 & 0.0795 & 0.0774 & 0.0794 & 0.0867 & 1.2326 & 1.1891 & \textbf{0.0749} \\
			& 3 & -      & 0.2750 & 0.2223 & 0.2586 & 0.2106 & 0.2103 & 0.2101 & 0.2254 & 1.2228 & 1.2959 & \textbf{0.1765} \\
			& 6 & -      & 0.4258 & 0.3723 & 0.3875 & 0.2976 & 0.3296 & 0.2801 & 0.3984 & 1.2239 & 1.2858 & \textbf{0.2770} \\ \hline
			\multirow{3}{*}{9}  & 1 & 0.0787 & 0.1996 & 0.0978 & 0.1138 & 0.0868 & 0.0806 & 0.0884 & 0.0995 & 1.2286 & 0.9772 & \textbf{0.0696} \\
			& 3 & -      & 0.3512 & 0.2183 & 0.2523 & 0.2171 & 0.1965 & 0.2112 & 0.2512 & 1.2264 & 1.2056 & \textbf{0.1692} \\
			& 6 & -      & 0.4942 & 0.3330 & 0.3743 & 0.2982 & 0.4093 & 0.3081 & 0.3422 & 1.2231 & 3.3885 & \textbf{0.2542} \\ \hline
			\multirow{3}{*}{12} & 1 & 0.0778 & 0.2113 & 0.0844 & 0.0916 & 0.0878 & 0.0799 & 0.0840 & 0.0930 & 1.2299 & 1.2733 & \textbf{0.0691} \\
			& 3 & -      & 0.3625 & 0.1839 & 0.1985 & 0.2116 & 0.1944 & 0.2099 & 0.2706 & 1.2282 & 1.4313 & \textbf{0.1692} \\
			& 6 & -      & 0.4880 & 0.2941 & 0.3239 & 0.2878 & 0.3850 & 0.3209 & 0.3262 & 1.2085 & 1.6453 & \textbf{0.2695} \\ \hline
		\end{tabular}
	}
	\end{center}
	\begin{tablenotes}
		\small
		\item {$^{\rm *}$ The p and q refers to the time lag and the predictive window size, respectively.}
	\end{tablenotes}
\end{table*}

To perform a comprehensive comparison, we perform three series of experiments.
We set the the predictive window size  to 1 to do a short period prediction,
and set the value to be 3 and 6  respectively to test the predictive ability in a longer period.
The loss functions of all the models average the predictive MSEs of the future weeks.
To evaluate the robustness of the model,
we also test each algorithm  with a time lag  of 6, 9, and 12, respectively. The value of $\lambda$ in this group of experiments is 0.01.

Table \ref{table:result-1} presents the results of all the models.
The results show that  the proposed DVGSN significantly outperforms all the baseline methods in all the prediction tasks with the predictive window size being 1, 3 and 6 respectively, which proves that DVGSN can satisfy both the short and long period prediction tasks.

\textbf{Time lag.} DVGSN shows a slight advantage with the time lag being 9. The results shows that  it is better enough to construct the virtual graph node and reflect the current trend with the recent 9 historical data. A longer time lag may offer no help  since the virtual graph  can learn the similar historical situation by itself. The result also shows that the selection of the hyperparameter time lag for DVGSN is relatively easy.

The other baseline methods, including KNN, RF, XG, MLP, and TPA-LSTM shows a better performance with a short time lag 6 for the short period prediction as the predictive window size is 1, while a bigger time lag 12 is better for a longer period prediction task in which the predictive window size is 3. A longer time lag could offer more support for the longer prediction task. However, as we have introduced previously, the time lag is limited and a longer one can reduce the training space.

The situation of TCN, TFR, GAT and GIN is more like a comprise  between the above two cases. They show  a better performance with a short time lag 6 for the short period prediction while a slight advantage with the time lag being 9 for the longer prediction task.  The graph-based solutions could reduce their dependence on the time lag. At the same time, the fact that they cannot benefit  from a longer time lag for the short prediction may be caused by their fixed static graph mode.

\section{Ablation Study}

To verify the effectiveness of the constructed dynamic graph, we designed a variant, denoted as ``DVGSN(fixed)", in which the virtual edges are fixed instead of being learned.
A given node is connected to the  nodes at the timepoints one week ago and one year (52 weeks) ago, considering the periodicity and time series of influenza.
Other preprocesses are the same as those in the proposed method.
We aslo  adjust the hyperparameter $\lambda$ to demonstrate the effectiveness of the penalty term in the loss function.

\begin{table*}[t]
	\caption{Comparison between the fixed and dynamic graph. The bold font indicates the better performance in each pair of comparison.}
	\label{table:result-fixed}
  \centering
	\resizebox{0.75\textwidth}{!}{
		\begin{tabular}{|c|c|c|c|c|c|c|c|c|c|}
			\hline
			\rowcolor[HTML]{c7c9cc}
			q                                                       & \multicolumn{3}{c|}{\cellcolor[HTML]{c7c9cc}1}      & \multicolumn{3}{c|}{\cellcolor[HTML]{c7c9cc}3}      & \multicolumn{3}{c|}{\cellcolor[HTML]{c7c9cc}6}      \\ \hline
			\rowcolor[HTML]{c7c9cc}
			p                                                       & 6               & 9               & 12              & 6               & 9               & 12              & 6               & 9               & 12              \\ \hline
			\begin{tabular}[c]{@{}c@{}}DVGSN \\ (fixed)\end{tabular}  & \textbf{0.0712} & 0.0773          & 0.0709          & 0.1817          & 0.1749          & 0.1780          & 0.3043          & 0.2699          & 0.2749          \\ \hline
			\begin{tabular}[c]{@{}c@{}}DVGSN \\ ($\lambda=0.01$)\end{tabular} & 0.0749          & \textbf{0.0696} & \textbf{0.0691} & \textbf{0.1765} & \textbf{0.1692} & \textbf{0.1692} & \textbf{0.2770} & \textbf{0.2542} & \textbf{0.2695} \\ \hline
		\end{tabular}
	}
\end{table*}

\begin{table*}[t]
	\caption{Comparison among different $\lambda$ in DVGSN. The bold font indicates the better performance in each pair of comparison.}
	\label{table:result-lambda}
\resizebox{1\textwidth}{!}{%
		\begin{tabular}{|c|c|c|c|c|c|c|c|c|c|c|c|}
			\hline
			\rowcolor[HTML]{c7c9cc}
			q & p  &
			$\lambda = 0$      &
			$\lambda = 0.0001$ &
			$\lambda = 0.0005$ &
			$\lambda = 0.001$  &
			$\lambda = 0.005$  &
			$\lambda = 0.01$   &
			$\lambda = 0.05$   &
			$\lambda = 0.1$    &
			$\lambda = 0.5$    &
			$\lambda = 1$
			\\ \hline
			1 & 6  & 0.0734 & 0.0763 & 0.0800 & 0.0700 & \textbf{0.0696} & 0.0749 & 0.0714 & 0.0705 & 0.0765 & 0.0710 \\ \hline
			1 & 9  & 0.0699 & \textbf{0.0657} & 0.0792 & 0.0751 & 0.0713 & 0.0696 & 0.0710 & 0.0735 & 0.0685 & 0.0698 \\ \hline
			1 & 12 & \textbf{0.0676} & 0.0898 & 0.0709 & 0.0712 & 0.0684 & 0.0691 & 0.0792 & 0.0759 & 0.0771 & 0.0775 \\ \hline
			3 & 6  & 0.1900 & 0.1874 & 0.1804 & 0.1875 & 0.1795 & \textbf{0.1765} & 0.1790 & 0.1772 & 0.1789 & 0.1888 \\ \hline
			3 & 9  & 0.1742 & 0.1777 & 0.1752 & 0.1779 & \textbf{0.1672} & 0.1692 & 0.1771 & 0.1739 & 0.1750 & 0.1708 \\ \hline
			3 & 12 & 0.1750 & 0.1873 & 0.1787 & 0.1699 & 0.1788 & \textbf{0.1692} & 0.1760 & 0.1727 & 0.1758 & 0.1777 \\ \hline
			6 & 6  & 0.2777 & 0.2758 & \textbf{0.2745} & 0.2964 & 0.2797 & 0.2770 & 0.2824 & 0.2765 & 0.3035 & 0.3082 \\ \hline
			6 & 9  & 0.3330 & 0.2797 & 0.2729 & 0.2626 & 0.2664 & \textbf{0.2542} & 0.2759 & 0.2767 & 0.2746 & 0.2962 \\ \hline
			6 & 12 & 0.2381 & \textbf{0.2376} & 0.2702 & 0.2531 & 0.2507 & 0.2695 & 0.2509 & 0.2675 & 0.2702 & 0.2618 \\ \hline
		\end{tabular}
	}
\end{table*}

\begin{figure}[!htb]
	\centering
	\includegraphics[width=1\linewidth]{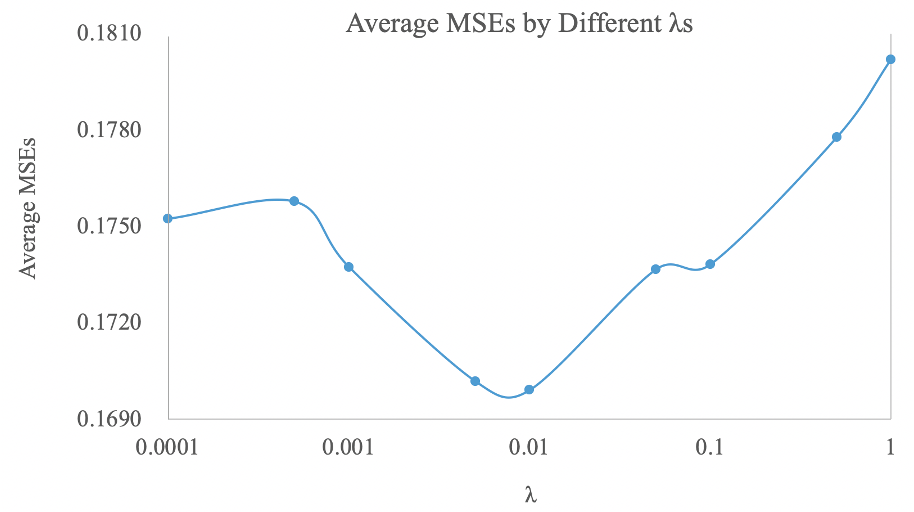}
	\caption{
		The average MSEs by the different $\lambda$s.
	}
	\label{figure:lambda}
\end{figure}

\textbf{Comparison between the fixed and dynamic graph.}
Table \ref{table:result-fixed} compares the performance of the fixed and dynamic graph.
In 8 of 9 cases, the dynamic graphs perform better.

\textbf{Comparison among different $\lambda$s in DVGSN.}
 A bigger $\lambda$ exerts a heavier penalty on a complex  virtual graphs with dense edges  of small  weights.
Table \ref{table:result-lambda} shows comparison among different $\lambda$s in DVGSN.
In the most cases, the ``DVGSN ($\lambda \neq 0$)" outperforms the ``DVGSN ($\lambda = 0$)".
In conclusion, restricting the complexity of the virtual graph improves the performance.

Figure \ref{figure:lambda} compares the average MSEs by the different $\lambda$s.
The X-axis and Y-axis represents $\lambda$s in logarithm scale and the column-average MSEs from Table \ref{table:result-lambda}, respectively.
The curve roughly presents a shape of the letter of ``V".
That is probably
because the penalty term cannot help to improve the robustness when the $\lambda$ is too small (close to zero).  When the $\lambda$ is too big, the loss will focus on the penalty term  but ignore the predictive MSE loss.
In a word, in this work, when the $\lambda$ approximately equals to 0.01, DVGSN performs best.


\begin{figure*}[!htb]
	\centering
	\includegraphics[width=.85\linewidth]{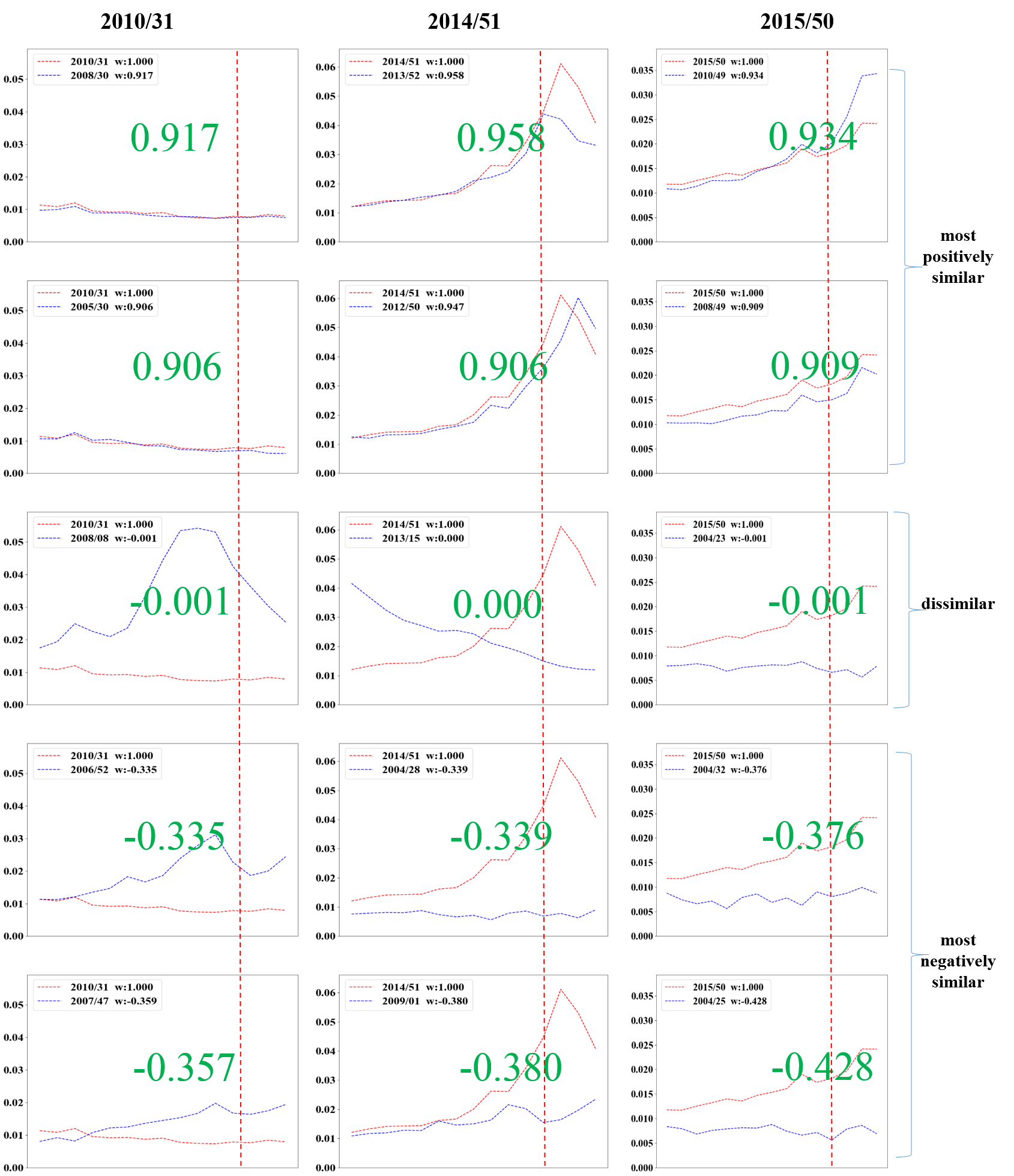}
	\caption{
		The illustrative examples of the positive similarities, dissimilarities, and negative similarities.
		The X-axis represents from the past 11th week ($p=12$) to the future 3rd week ($q=3$) to predict.
		The red dash lines separate past(including current) and future timepoints.
		The Y-axis represents the ILI rate.
		The date above each column of the subfigures is the given timepoint.
		The green float in each subfigure is the significance of the similarity between the given virtual node (the red curve) and the neighbor node (the blue curve).
	}
	\label{figure:learned_effect}
\end{figure*}

\begin{figure}[!htb]
	\centering
	\includegraphics[width=1\linewidth]{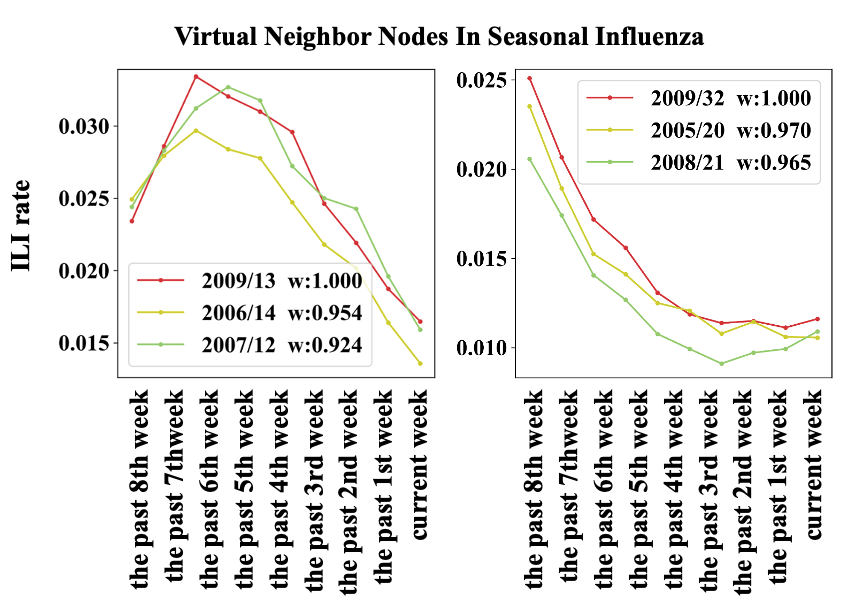}
	\caption{
		The illustrative examples how  DVGSN deals with the varied seasonality.
		The X-axis represents from the past 9th week to the upcoming 3rd week ($p=9$ and $q=3$).
		The ``infection situations" of the given timepoints are represented by the red curves; and
		the most two similar ``infection situations" are represented by the blue and green curves.
		The number after ``w:"  in the figure legend is the significance of the similarity.
	}
	\label{figure:varied_season}
\end{figure}

\begin{figure}[!htb]
	\centering
	\includegraphics[width=1\linewidth]{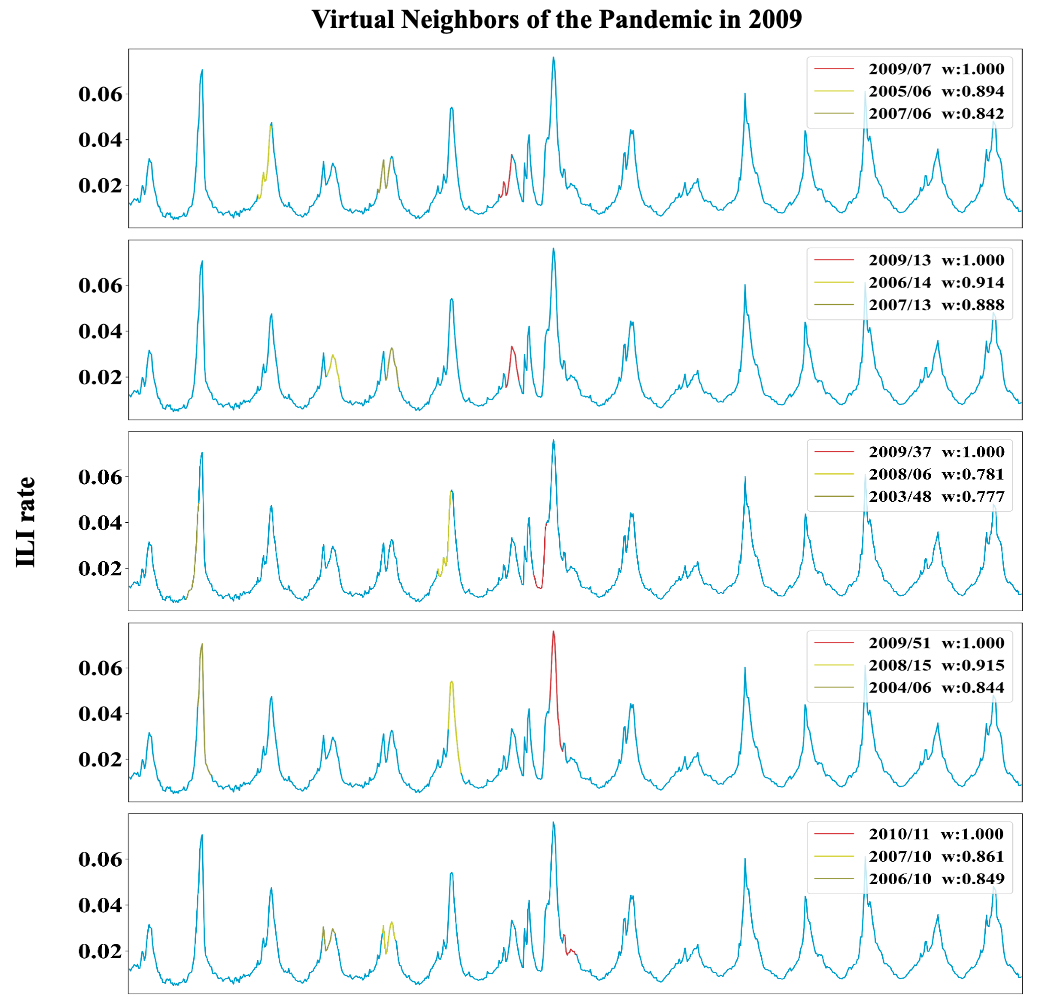}
	\caption{
		This figure illustrates  how  DVGSN learns from the historical ``infection situations" and predicts the pandemic.
		 The X-axis represents time series from 2002/40 to 2017/30.
	}
	\label{figure:pandemic}
\end{figure}

\section{Model Interpretation}

This section explains how the proposed method works.

\subsection{How does DVGSN learn the significance of the similarity?}

Figure \ref{figure:learned_effect} illustrates the similarity that the virtual graph learns.
The X-axis represents the time series from the past 11th week to the future 3rd week ($p=12$ and $q=3$).
The Y-axis represents the ILI rate.
The date in the format of ``year/week" above each column of the subfigures is the ``current" timepoint.
The model predicts the ILI rates of the future 3 weeks (on the right side of the red dash line).
In each subfigure, the red curves represent the ``infection situations" of the given timepoints.
The ILI rates of the current and past 11 weeks (on the left side of the red dash line) are projected to a high-dimensional space to calculate the significance of the similarity (the green float in each subfigure).
As a result,
the blue curves in the top two subfigures are the most positively similar ``infection situations" that the virtual graph finds;
the blue curves in the third row of the subfigure represent dissimilar ``infection situations" that the virtual graph finds; and
the blue curves in the bottom two subfigures are the most negatively similar ``infection situations" that the virtual graph finds.

\subsection{How does  DVGSN learn for the varied seasonality?}

This section explores whether DVGSN can deal with the varied influenza seasonality.
Figure \ref{figure:varied_season} gives two examples.
The X-axis represents the time series from the past 9th week to the upcoming 3rd week ($p=9$ and $q=3$).
The red curves represent the ``infection situations"  of the given timepoints; and
the blue and green curves represent the two most similar ``infection situations".
We find the two most similar timepoints do not correspond to the same week of the previous years, which demonstrates DVGSN can tackle varied influenza seasonality instead of sticking to the periodicity of 52 weeks.

\subsection{How does  DVGSN learn for the pandemic?}

This section explores how DVGSN learns from the historical ``infection situations" and predicts the pandemic.
We present five examples in the 2009 pandemic in Figure \ref{figure:pandemic}.
The X-axis and Y-axis represents time series from 2002/40 to 2017/30 and the ILI rates, respectively.
In each subfigure, the red piece represents the ``infection situations"  of the given timepoints in the pandemic.
The timepoints in the five subfigures is 2009/07, 2009/13, 2009/37, 2009/51, and 2010/11 respectively, which represent a rising, a falling down, a rebound after reaching a bottom, a drop after reaching a peak, and fluctuations after a huge dropping in the 2009 pandemic respectively.
The two yellow pieces in the curves represent two of the most similar ``infection situations"  that DVGSN learns.
The number after ``w:"  in each figure legend is the significance of the similarity.
By comparing the past ``infection situations"  and future tendency between the red pieces and the two yellow pieces in the five examples, we conclude that DVGSN can find and learn the similar ``infection situations"  outside the pandemic and thereby make a reliable model for the influenza prediction.

\section{Conclusion}

In this work, we proposed a method---DVGSN.
DVGSN can find similar ``infection situations" outside the time window and therefore improve the predictive accuracy for influenza.
The extensive experiments on real-world influenza data demonstrate that DVGSN significantly outperforms the current state-of-the-art methods.
Besides, the proposed method has rich interpretabilities, which provide us clues how the model perform prediction for influenza. Another strong point of the  proposed method lies in that it  need  less domain knowledge to build a graph in advance, which may be very difficult in the medical science related fields. As all the deep learning based methods the proposed method also depends on enough data to train the model.
Hopefully, this method can help us better prepare for influenza outbreaks, and work on other public health related analytical tasks well.


%

\appendices
%

\ifCLASSOPTIONcompsoc
  \section*{Acknowledgments}
  This work was supported   by the Strategic Priority Research Program of Chinese
  Academy of Sciences (Grant No. XDB38040200). We would like to thank all the authors of the open source code in the baseline methods.
\else
  \section*{Acknowledgment}
\fi

\ifCLASSOPTIONcaptionsoff
  \newpage
\fi



\bibliographystyle{IEEEtran}
\bibliography{IEEEabrv,mybibfile.bib}
\end{document}